\documentclass{article}

\usepackage[preprint]{neurips_2026}

\usepackage[utf8]{inputenc} % allow utf-8 input
\usepackage[T1]{fontenc}    % use 8-bit T1 fonts
\usepackage{hyperref}       % hyperlinks
\usepackage{url}            % simple URL typesetting
\usepackage{booktabs}       % professional-quality tables
\usepackage{amsfonts}       % blackboard math symbols
\usepackage{nicefrac}       % compact symbols for 1/2, etc.
\usepackage{microtype}      % microtypography
\usepackage{xcolor}         % colors

\usepackage{amsmath}
\usepackage{amssymb}
\usepackage{mathtools}
\usepackage{amsthm}

\usepackage[dvipsnames]{xcolor}         % colors

\usepackage{booktabs}       % professional-quality tables
\usepackage{amsfonts}       % blackboard math symbols
\usepackage{nicefrac}       % compact symbols for 1/2, etc.
\usepackage{microtype}      % microtypography
\usepackage{colortbl}

\usepackage{MnSymbol,bbding,pifont}
\usepackage{wrapfig}
\usepackage{enumitem}
\usepackage{caption}
\usepackage{tikz,relsize,amsmath,booktabs}
\usepackage{array} 
\usepackage{subcaption}
\usepackage{xspace}
\usepackage{multirow}
\usepackage{makecell}
\usepackage{tablefootnote}
\usepackage{graphicx}
\usepackage{boxedminipage}
\usepackage{listings}
\usepackage{threeparttable}
\usepackage{tcolorbox}
\usepackage{alltt}
\usepackage{algorithm}
\usepackage{algorithmic}  
\usepackage{xcolor}
\usepackage{colortbl}

\definecolor{darkgreen}{RGB}{50,100,0}
\definecolor{darkred}{RGB}{200, 0, 0}
\definecolor{firstBest}{rgb}{0.86, 1, 0.86}
\definecolor{secondBest}{rgb}{1, 0.91, 0.93}

\newcommand{\cxmark}{\ding{52}\rotatebox[origin=c]{-9.2}{\kern-0.7em\ding{55}}}

\newcommand{\ie}{\emph{i.e.,}\xspace}

\newcommand{\etc}{\emph{etc.}\xspace}

\newcommand{\method}{\textsc{Hint Tuning}\xspace}

\theoremstyle{plain}

\theoremstyle{definition}

\theoremstyle{remark}

\title{Hint Tuning: Less Data Makes Better Reasoners}

\author{
  Siqi Fan$^{1}$\quad Minghao Li$^{2}$\quad Xiaoqian Ma$^{2}$\quad Xiusheng Huang$^{2}$\quad Zhuo Chen$^{2}$ \quad \\ \textbf{Bowen Qin}$^{3}$ \quad \textbf{Liujie Zhang}$^{2}$ \quad \textbf{Shuo Shang}$^{1}$ \quad \textbf{Weihang Chen}$^{2}$\\
  $^1$University of Electronic Science and Technology of China\\ $^2$Xiaohongshu Inc.  \\ 
  $^3$National University of Singapore
}

\begin{document}

\maketitle

\begin{abstract}
Large reasoning models achieve high accuracy through extended chain-of-thought but generate 5--8$\times$ more tokens than necessary, applying verbose reasoning uniformly regardless of problem difficulty. We propose \textbf{\method}, a data-efficient approach that teaches models to calibrate reasoning depth. Our key insight: the \textbf{corresponding instruct model} serves as an ideal difficulty probe. By testing what the instruct model can solve with varying guidance, we automatically construct training data across three states: \textit{No-Hint} (direct answer), \textit{Sparse-Hint} (minimal prefix), and \textit{Full-Hint} (complete reasoning). This converts the abstract challenge of difficulty labeling into a measurable consistency check between the instruct and reasoning models.

With only {1K self-annotated samples}, \method achieves {24--66\% token reduction} (31.5\% average) across mainstream reasoning models (Qwen3-Thinking, DeepSeek-R1-Distill) at multiple scales (4B--32B) while maintaining competitive accuracy on five benchmarks. Unlike methods requiring massive distillation datasets or expensive RL, we achieve superior efficiency through simple alignment with the instruct model's capabilities. Code and data are available at \url{https://github.com/redai-infra/hint-tuning}.
\end{abstract}

\begin{wrapfigure}[13]{r}{0.5\linewidth}
\vspace{-1em}
\centering
\includegraphics[width=\linewidth]{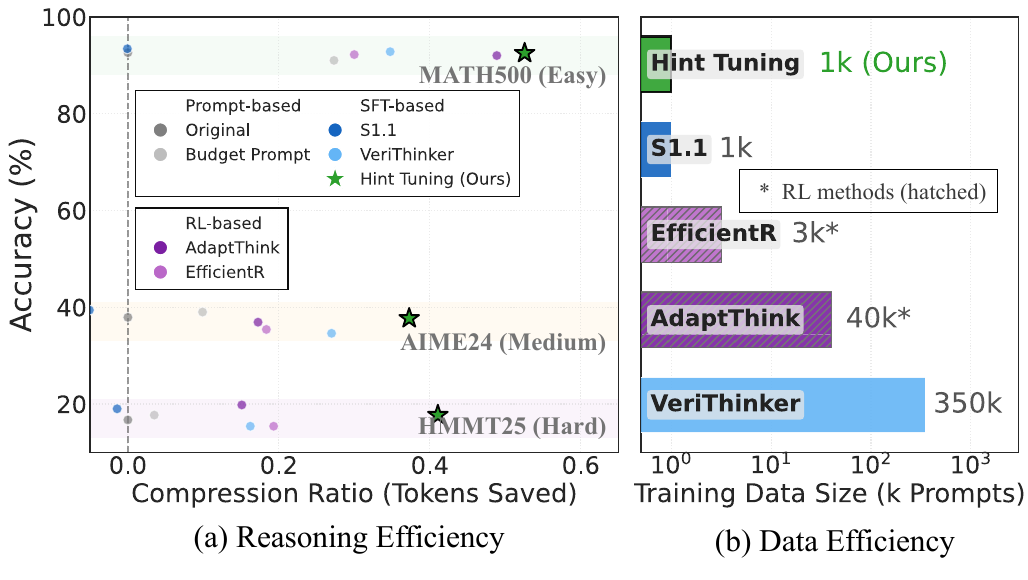}
\caption{\method Efficiency on DeepSeek-R1-Distill-Qwen-7B.}
\label{fig:efficiency_comparison}
\end{wrapfigure}

\section{Introduction}
\label{sec:intro}

Chain-of-thought (CoT) prompting~\citep{wei2022chain,DBLP:conf/nips/KojimaGRMI22} has become the dominant paradigm for eliciting reasoning in large language models. Recent work demonstrates that CoT follows test-time scaling laws~\citep{snell2024scaling,openai2024learning}: accuracy improves with longer reasoning traces, enabling breakthroughs on complex mathematical tasks~\citep{openai2024o1,guo2025deepseek}. However, this capability introduces a new inefficiency: \textbf{models reason verbosely on all problems}, regardless of difficulty.

\begin{figure}[t]
  \centering
  \includegraphics[width=\linewidth]{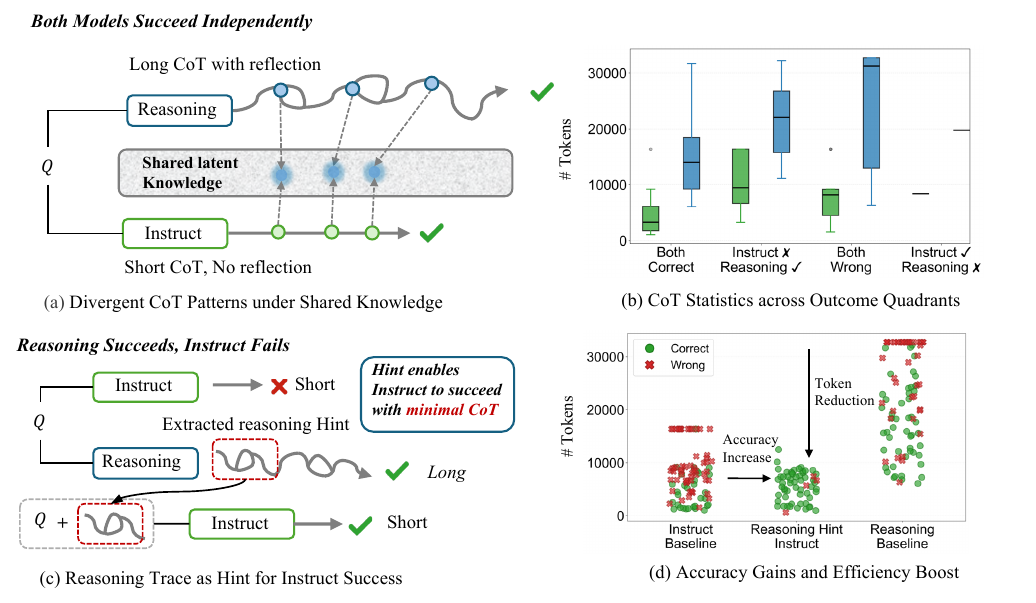}
  \caption{{Reasoning models waste tokens through over-elaboration.}
  (a) Example: same answer, different verbosity.
  (b) Systematic comparison: instruct models use $5$--$8\times$ fewer tokens across all four correctness scenarios.
  (c--d) With partial hints, instruct models need only 27.5\% of episodes and 11.9\% of tokens to succeed.}
  \label{fig:reasoning_hint}
\end{figure}

As shown in Figure~\ref{fig:reasoning_hint}, comparing an instruct and a reasoning model across four correctness scenarios reveals that the instruct model consistently uses far fewer tokens. Even when both succeed, the reasoning model generates $5$--$8\times$ more tokens through verbose self-reflection~\citep{chen2024not,sui2025stop}, wasting computation without adding reasoning value.

\paragraph{The annotation bottleneck.}
Training models to adapt reasoning depth requires supervision signals specifying ``how much thinking is appropriate.'' Existing approaches struggle with this challenge in complementary ways. \emph{Post-hoc compression} methods~\citep{kang2025c3ot,DBLP:journals/corr/abs-2502-12067,chen2025verithinker} prune redundant steps via heuristics or external models, but importance is context-dependent: the same phrase may be essential in one problem and redundant in another. \emph{During-reasoning} methods~\citep{han2025token,liu2024can,ma2025cotvalve} generate concise reasoning via token budgets or step-skipping, though uniform strategies fail on problems of varying complexity. Both families avoid the cost of \emph{RL with length penalties}~\citep{luo2025o1pruner,arora2025training,aggarwal2025l1,zhang-etal-2025-adaptthink}, which requires careful tuning of penalty coefficients and reward scaling, incurring significantly higher compute than SFT~\citep{sui2025stop,schulman2017}; moreover, fixed penalties cause over-compression on easy problems and under-compression on hard ones.

\paragraph{The instruct model as a capability probe.}
Our key insight is that the {corresponding instruct model} serves as an ideal capability probe. If the instruct model can solve a problem with minimal or no reasoning guidance, the reasoning model should not generate a lengthy chain-of-thought. By probing the instruct model with incrementally more reasoning \textit{episodes}\footnote{Following~\citep{qu2025optimizing}, each round of verification is an \textit{episode}; we use regex patterns such as ``let me verify'' or ``on second thought'' to segment episodes.} from the reasoning model's trace, we categorize problems into three cognitive states. \textbf{State~I (No-Hint):} the instruct model succeeds without any hint, so the reasoning model should produce a direct answer. \textbf{State~II (Sparse-Hint):} the instruct model succeeds given only the first few episodes---the ``Minimum Effective Hint''---so the reasoning model should generate just this essential prefix. \textbf{State~III (Full-Hint):} the instruct model fails even with substantial hints, so the reasoning model should generate complete deliberative reasoning to ensure accuracy.

\paragraph{\method: efficiency via alignment.}
Based on this self-annotation, we propose {\method}. Unlike RL methods that must \emph{discover} efficiency through expensive exploration, we frame it as an \emph{alignment} task: aligning the reasoning model's output length with the instruct model's capability boundaries. This converts the ill-defined question of ``how much reasoning is enough'' into concrete, measurable training data, requiring no human annotation or reward models.

\paragraph{Contributions.}
On Qwen3 and DeepSeek-R1-Distill models (4B--32B), \method achieves {24--66\% token reduction} (31.5\% on average) with competitive accuracy across five math reasoning benchmarks (Figure~\ref{fig:efficiency_comparison}a), using only {1K training samples} (25--640$\times$ fewer than existing methods, Figure~\ref{fig:efficiency_comparison}b).

\section{Related Works}
\label{sec:related}
\begin{wrapfigure}{r}{0.5\linewidth}
\vspace{-1.5em}
  \centering
  \includegraphics[width=\linewidth]{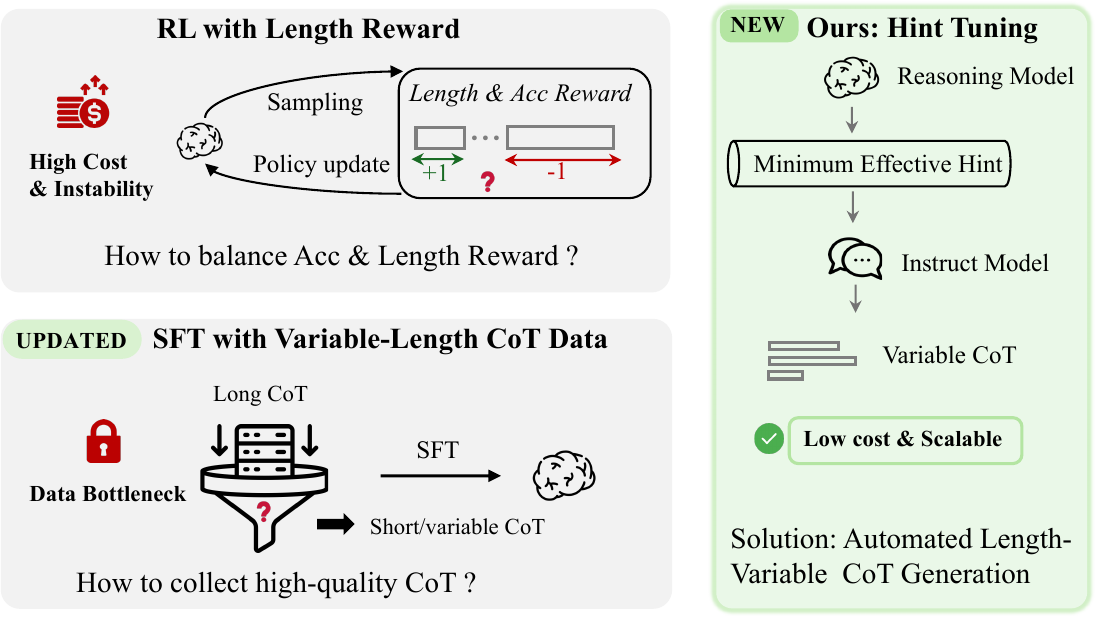}
\caption{Efficient reasoning training paradigms.}
    \label{fig:method_landscape}
\end{wrapfigure}
\paragraph{Chain-of-Thought and Test-Time Scaling.}
Chain-of-Thought (CoT) prompting~\citep{wei2022chain, DBLP:conf/nips/KojimaGRMI22} elicits reasoning in large language models by encouraging step-by-step thinking. Recent work shows CoT follows a test-time scaling law~\citep{snell2024scaling,openai2024learning}, where accuracy improves with increased inference compute via \textit{sequential scaling} (longer reasoning traces per sample, as in O1 and DeepSeek-R1) or \textit{parallel scaling} (aggregating multiple samples via self-consistency~\citep{DBLP:conf/iclr/0002WSLCNCZ23} or best-of-N)~\citep{muennighoff2025s1}. However, this improved accuracy comes at the cost of substantial computational overhead and increased latency.

\paragraph{SFT-Based Efficient Reasoning.}
One line of work improves reasoning efficiency via supervised fine-tuning. Existing approaches employ two strategies: \textit{post-reasoning compression} prunes redundant steps from full traces using heuristics or external models~\citep{kang2025c3ot, DBLP:journals/corr/abs-2502-12067, chen2025verithinker, DBLP:conf/acl/MunkhbatHKYKY25}, while \textit{during-reasoning generation} prompts models to produce concise paths through optimal token budgets~\citep{han2025token}, step-skipping~\citep{liu2024can}, or parameter mixing~\citep{ma2025cotvalve}.
Fine-tuning strategies include standard approaches (LoRA~\citep{hu2022lora}, DPO~\citep{DBLP:conf/acl/HanWFZM025,DBLP:conf/nips/RafailovSMMEF23}), progressive methods~\citep{ma2025cotvalve, DBLP:conf/nips/LiuGHJZQZ24}, and model merging~\citep{team2025kimi}.
However, these methods determine reasoning length based on external models or uniform strategies, ignoring that problem difficulty is relative to the target model's capability-what appears simple to a strong compressor may still require extensive reasoning for the model being fine-tuned, causing inappropriate compression.

\paragraph{RL-Based Efficient Reasoning.}
Reinforcement Learning (RL) incorporate length penalties into reward functions via cosine-based penalties~\citep{yeo2025demystifying}, length-harmonizing rewards~\citep{luo2025o1pruner, arora2025training, feng2026pace}, L1 regularization~\citep{aggarwal2025l1}, constrained instructions~\citep{shen2025dast}, or preference optimization~\citep{zhang-etal-2025-adaptthink, DBLP:conf/nips/RafailovSMMEF23}. These methods integrate length-aware objectives into classical RLHF frameworks such as PPO~\citep{schulman2017} and DPO~\citep{DBLP:conf/nips/RafailovSMMEF23, ouyang2022training}. However, RL incurs higher compute costs than SFT with training instability~\citep{zheng2023secrets}. Fixed penalties fail to adapt to problem heterogeneity~\citep{cobbe2021training, lightman2023let}, causing over-compression on easy tasks and under-compression on hard ones. Despite automation claims, extensive hyperparameter tuning remains necessary~\citep{engstrom2020implementation, andrychowicz2020matters}: PPO clips, KL constraints, reward scaling, and sampling temperatures.

\paragraph{Positioning of Our Work.}
As illustrated in Figure~\ref{fig:method_landscape}: instead of expensive RL optimization or data-limited SFT, we automatically generate difficulty-adaptive reasoning traces via \textit{Minimum Effective Hint} criterion. This data-centric strategy achieves RL-level efficiency at lower cost, requiring no reward models or external compressor models for data labeling.
\section{Method}
\label{sec:method}

Reasoning Models generate verbose traces regardless of problem difficulty, wasting computation on problems that require far less reasoning. We first quantify this inefficiency, then describe how it motivates our data construction pipeline.

\subsection{Efficiency Gap and Capability Probing}
\label{subsec:probing}

We compare Qwen3-4B-Instruct (Instruct Model) and Qwen3-4B-Thinking (Reasoning Model)~\citep{qwen3technicalreport} on three competition-level math benchmarks: AIME24, AIME25, and HMMT25, sampling 8 responses per problem validated with \texttt{Math-Verify}. As shown in Figure~\ref{fig:reasoning_hint}, Reasoning Models generate $5$--$8\times$ more tokens than Instruct Models even when both reach correct answers. Yet when provided with partial reasoning hints, Instruct Models successfully solve problems using only 27.5\% of episodes and 11.9\% of tokens, suggesting most reasoning is redundant.

This motivates a key question: \textit{what is the minimum hint needed to activate the Instruct Model's capabilities?} We define the \textbf{Minimum Effective Hint} $K^*$ as:
\begin{equation}
K^* = \min\bigl\{k \in \{0, 1, \ldots, N\} : M_{\text{instruct}}(x \oplus e_{1:k}) = y\bigr\}
\label{eq:mec}
\end{equation}
where $e_{1:k}$ denotes the first $k$ reasoning episodes from the Reasoning Model ($e_{1:0} = \emptyset$). If no such $k$ exists (\ie the instruct model fails under all hint levels), we set $K^* = N$, treating the problem as requiring full deliberation.
Intuitively, $K^*$ measures how much of the Reasoning Model's trace is actually necessary to unlock the Instruct Model's answer.

\subsection{Hint Distribution Analysis}
\label{subsec:hint_distribution}

\begin{figure}[t]
  \centering
  \includegraphics[width=\linewidth]{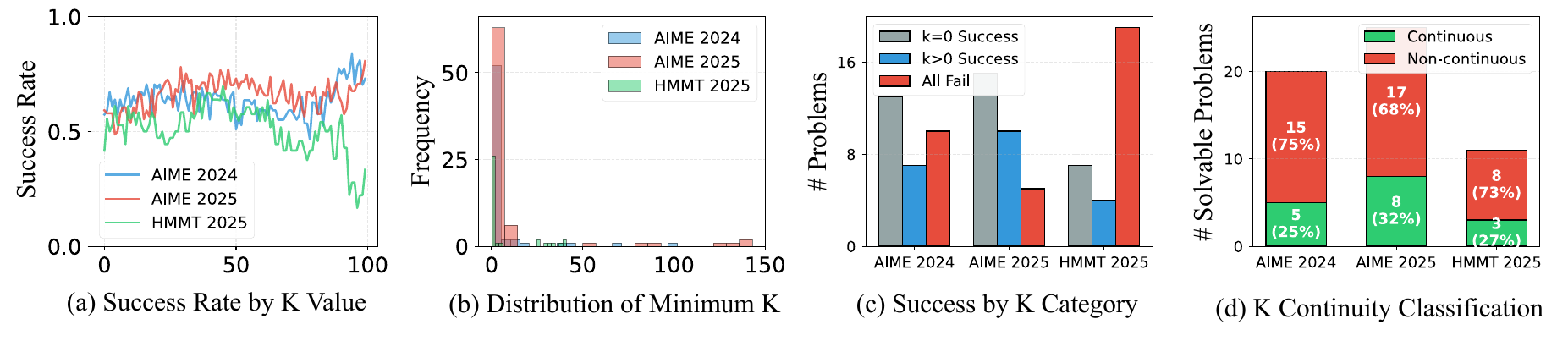}
  \caption{\textbf{Distribution of $K^*$ reveals non-monotonic patterns.}
  (a) Success rate fluctuates with $K$ rather than increasing monotonically.
  (b) Most required hints concentrate in early episodes (within first 25).
  (c) Hints enable 23--33\% of previously unsolvable problems to succeed with sparse reasoning, and 33--60\% with full traces.
  (d) Only 25--32\% show continuous success; most exhibit non-monotonic patterns.}
  \label{fig:kmin_distribution}
\end{figure}

Analyzing the distribution of $K^*$ (Figure~\ref{fig:kmin_distribution}) reveals three properties that challenge the ``longer is better'' assumption.

\paragraph{Hints unlock previously unsolvable problems.}
While 36--50\% of problems require no hints at all, reasoning hints enable an additional 23--33\% to succeed with sparse episodes and 33--60\% with full traces (Figure~\ref{fig:kmin_distribution}c), demonstrating that the Instruct Model has latent capabilities that targeted hints can activate.

\paragraph{Key steps matter more than length.}
Only 25--32\% of problems exhibit \textit{continuous} success, where succeeding at episode $k$ guarantees success at $k{+}1$. For the remaining 68--75\%, success is non-monotonic: additional episodes can introduce distractors, and success rates fluctuate rather than consistently improve with $K$ (Figure~\ref{fig:kmin_distribution}a,d). This indicates that specific pivotal episodes matter far more than linear accumulation.

\paragraph{Early hints serve as activators.}
Critical reasoning steps concentrate within the first 25 episodes (Figure~\ref{fig:kmin_distribution}b), suggesting CoT hints function as direction-setting activators rather than scaffolding that requires complete reasoning chains.

\subsection{Constructing Variable-Length CoT Data}
\label{subsec:pipeline}

The three properties above motivate a natural three-state partition. We use the s1 dataset~\citep{muennighoff2025s1} as our seed corpus---1,000 challenging problems curated from 59K samples across 16 diverse sources (NuminaMath, OlympicArena, \etc), filtered for quality and difficulty. The pipeline transforms these problems into efficient training samples through three stages: (1)~\textbf{Generation}: we discard the original s1K traces and regenerate fresh CoT traces for each problem $x$ using $M_{\text{reason}}$; (2)~\textbf{Segmentation}: we decompose traces into discrete episodes $(e_1, \ldots, e_N)$ based on self-reflection markers (Appendix~\ref{app:segmentation}); (3)~\textbf{Probing}: we test $M_{\text{instruct}}$ by sequentially prepending $k$ episodes to determine if it reaches the correct answer $y$, recording $K^*$.

Based on $K^*$, we construct the training response $r$ for each problem:
\begin{equation}
    r = \texttt{<think>} \oplus 
    \begin{cases}
        p_0 & K^* = 0 \\
        (e_1, \ldots, e_{K^*}) & 0 < K^* < N \\
        (e_1, \ldots, e_N) & K^* = N
    \end{cases}
    \oplus \; \texttt{</think>} \oplus y
    \label{eq:response}
\end{equation}
where $p_0$ is a brief directive (e.g., ``Direct answer'') for \textbf{State~I} (No-Hint); the first $K^*$ episodes are retained for \textbf{State~II} (Sparse-Hint); and the full trace is preserved for \textbf{State~III} (Full-Hint). The complete procedure is formalized in Algorithm~\ref{alg:pipeline} (Appendix~\ref{app:algorithm}). The resulting dataset is balanced across the three states (35.7\% / 26.4\% / 38.2\%), reflecting diverse difficulty levels in the s1K corpus; see Appendix~\ref{app:data} for statistics and Appendix~\ref{app:cases} for representative examples.

We combine $\mathcal{D} = \mathcal{D}_1 \cup \mathcal{D}_2 \cup \mathcal{D}_3$ and fine-tune $M_{\text{reason}}$ via standard cross-entropy loss, teaching it to adapt output length to problem difficulty---generating direct answers for simple problems while preserving full deliberative reasoning for complex ones.
\begin{table}[t]
\centering
\small
\caption{Overview of evaluation benchmarks.}

\begin{tabular}{llrc}
\toprule
\textbf{Category} & \textbf{Benchmark} & \textbf{\# Prob.} & \textbf{Difficulty} \\
\midrule
\multirow{4}{*}{In-domain} 
    & MATH500 & 500 & High School \\
    & AIME24 & 30 & Competition \\
    & AIME25 & 30 & Competition \\
    & HMMT25 & 30 & Competition \\
\cmidrule(l){2-4}
Temporal Held-out & AMO-Bench & 50 & IMO-level \\
\midrule
\multirow{2}{*}{Cross-domain} 
    & LiveCodeBench & 713 & Mixed \\
    & GPQA-Diamond & 198 & Graduate \\
\bottomrule
\end{tabular}
\label{tab:benchmarks}
\end{table}

%==============================================================================
% SECTION 5: EXPERIMENTS
%==============================================================================
\section{Experiments}
\label{sec:exp}

We evaluate \method across four model scales and three task domains. Our experiments are designed to answer: (1) Can \method reduce reasoning verbosity without sacrificing accuracy? (2) Does efficiency generalize to unseen problems and domains? (3) What does adaptive compression reveal about the nature of reasoning itself?

%------------------------------------------------------------------------------
\subsection{Experimental Setup}
\label{subsec:setup}

\paragraph{Models.}
For reasoning models, we evaluate Qwen3-4B-Thinking~\citep{qwen3technicalreport} and DeepSeek-R1-Distill-Qwen-\{7B, 14B, 32B\}~\citep{guo2025deepseek}, spanning an order of magnitude in scale. We use Qwen3-4B-Instruct as the Instruct Model for hint validation.

\paragraph{Baselines.}
We compare against four categories of methods. \textbf{Original Models} report zero-shot results of vanilla reasoning models. \textbf{Prompt-based}: Budget Prompt~\citep{han2024token} instructs models to ``think within $N$ tokens'' at inference time without weight updates (budget set to 8,000; see Appendix~\ref{app:baseline}). \textbf{SFT-based}: SFT-S1.1 fine-tunes on high-quality traces from s1~\citep{muennighoff2025s1}; VeriThinker~\citep{chen2025verithinker} distills from self-verified reasoning paths to suppress unnecessary reflection (340K samples). \textbf{RL-based}: EfficientR~\citep{arora2025training} and AdaptThink~\citep{zhang-etal-2025-adaptthink} (7B only) apply online RL with length-accuracy reward shaping. EfficientR uses 3.2K prompts with 8 rollouts per prompt (25.6K total generations); AdaptThink uses 40K prompts with 16 rollouts per prompt (640K total generations). Both 
incur substantially higher compute than offline SFT methods.

\paragraph{Benchmarks.}
We evaluate across a spectrum of domain familiarity (Table~\ref{tab:benchmarks}). For {in-domain} math reasoning, we use MATH500~\citep{DBLP:conf/nips/HendrycksBKABTS21}, AIME24, AIME25, and HMMT25, spanning increasing difficulty and chronological release~\citep{cao2025toward}. As a {temporal held-out} set, AMO-Bench~\citep{an2025amobench} comprises 50 IMO-level problems released after all model checkpoints, minimizing contamination risk. For \textbf{cross-domain} generalization, we include LiveCodeBench~\citep{DBLP:conf/iclr/JainHGLYZWSSS25} (code) and GPQA-Diamond~\citep{rein2024gpqa} (science).

\paragraph{Evaluation Protocol.}
All evaluations use the \texttt{lighteval} framework\footnote{\url{https://huggingface.co/docs/lighteval}} for reproducibility. We sample 16 responses per problem with temperature 0.6 and top-$p$ 0.95, reporting averaged accuracy to mitigate variance. All problems use a consistent prompt: \textit{``Please reason step by step, and put your final answer within \textbackslash boxed\{\}.''} Our baseline reproductions align with official reports.

\subsection{Implementation Details}
\label{subsec:implementation}
We sample 1K problems from s1~\citep{muennighoff2025s1} and apply our three-state annotation strategy (Section~\ref{sec:method}). For each problem, we probe$M_{\text{instruct}}$ by prepending episode prefixes to determine $K^*$(Equation~\ref{eq:mec}). We use Qwen3-4B-Instruct as the default probe across all model scales to ensure a consistent and fair comparison. While a stronger same-family instruct model can improve label quality at larger scales where the probe-target capability gap is more pronounced, we find that the 4B probe already yields competitive results across most settings (Section~\ref{subsec:probe_selection}). 
For practitioners, we recommend using the strongest available same-family instruct model as the probe when accuracy is the primary concern. Since 90\% of hint-requiring problems succeed within the first 25 episodes, we apply early stopping during the linear search, keeping the average number of probes per problem to only 3--6. Samples are labeled as {No-Hint} ($K^*{=}0$), {Sparse-Hint} ($0 < K^* < N$), or {Full-Hint} ($K^*{=}N$).
We perform full-parameter SFT with learning rate $1 \times 10^{-5}$, cosine decay, and max sequence length 32,768. Training takes 2/2.5/8/10 hours for 4B/7B/14B/32B models on 8$\times$H20 GPUs. See Appendix~\ref{app:training} for complete details.

%------------------------------------------------------------------------------

\begin{table}[!t]
\centering
\caption{Main results across model scales and reasoning benchmarks. We compare against (1) \textit{General baselines} (Original, SFT, Budget Prompt) implemented at all scales, and (2) \textit{Specialized methods} (AdaptThink, Efficient Reasoning, VeriThinker) with official checkpoints available only at 7B scale. For \method, improvements over Original are shown in \textcolor{green!70!black}{green}, declines in \textcolor{red!70!black}{red}.}
\label{tab:main_results}
\resizebox{\linewidth}{!}{
\begin{tabular}{l|c|cc|cc|cc|cc|cc|cc}
\toprule
\multirow{3}{*}{\textbf{Method}} 
& \multirow{3}{*}{\textbf{\# Data}}
& \multicolumn{10}{c|}{\textbf{Mathematical Reasoning Tasks}} 
& \multicolumn{2}{c}{\textbf{All}} \\
\cmidrule(lr){3-12} \cmidrule(lr){13-14}

& 
& \multicolumn{8}{c|}{\textbf{Standard Benchmarks}} 
& \multicolumn{2}{c|}{\textbf{Temporal Gen.}}
& \multicolumn{2}{c}{\multirow{2}{*}{\textbf{Average}}} \\
\cmidrule(lr){3-4} \cmidrule(lr){5-6} \cmidrule(lr){7-8} \cmidrule(lr){9-10} \cmidrule(lr){11-12}

& 
& \multicolumn{2}{c|}{\textbf{MATH500}} 
& \multicolumn{2}{c|}{\textbf{AIME24}} 
& \multicolumn{2}{c|}{\textbf{AIME25}}
& \multicolumn{2}{c|}{\textbf{HMMT25}}
& \multicolumn{2}{c|}{\textbf{AMO-Bench}}
& \multicolumn{2}{c}{} \\
\cmidrule(lr){3-4} \cmidrule(lr){5-6} \cmidrule(lr){7-8} \cmidrule(lr){9-10} \cmidrule(lr){11-12} \cmidrule(lr){13-14}

& & Acc & Tok 
& Acc & Tok 
& Acc & Tok
& Acc & Tok
& Acc & Tok
& Acc & Tok \\
\midrule

% ================= Qwen 4B =================
\multicolumn{14}{c}{\cellcolor{gray!25}\textit{Qwen3-4B-Thinking-2507}} \\ 
\midrule
\color{gray}Original & \color{gray}-- &\color{gray}97.8&\color{gray}6517&\color{gray}77.9&\color{gray}19095&\color{gray}71.3&\color{gray}20973&\color{gray}39.4&\color{gray}24376&\color{gray}10.3&\color{gray}27533&\color{gray}59.3&\color{gray}19699 \\

Budget Prompt & -- & 97.2 & 5457 & 77.1 & 18147 & 71.0 & 20362 & 41.7 & 22767 & 12.8 & 25509 & 60.0 & 18448 \\
SFT-S1.1 & 1K  & 95.3 & 3250 & 67.1 & 10929 & 56.0 & 12273 & 30.8 & 14222 & 7.7 & 18881 & 51.4 & 11911 \\

\cmidrule[0.4pt](l){1-14}

\textbf{\method} & \textbf{1K}&
95.7 & 2190 & 73.8 & 10233 & 59.4 & 12099 & 32.7 & 14523 & 20.5 & 25926 & 56.4 & 12994 \\
\quad \textit{vs. Original} & -- & 
\textcolor{red!70!black}{-2.1} & \textcolor{green!70!black}{-66\%} & 
\textcolor{red!70!black}{-4.1} & \textcolor{green!70!black}{-46\%} & 
\textcolor{red!70!black}{-11.9} & \textcolor{green!70!black}{-42\%} & 
\textcolor{red!70!black}{-6.7} & \textcolor{green!70!black}{-40\%} & 
\textcolor{green!70!black}{+10.2} & \textcolor{green!70!black}{-6\%} &
\textcolor{red!70!black}{-2.9} & \textcolor{green!70!black}{-34\%} \\

\midrule

% ================= DeepSeek 7B =================
\multicolumn{14}{c}{\cellcolor{gray!25}\textit{DeepSeek-R1-Distill-Qwen-7B}} \\ 
\midrule
\multicolumn{14}{l}{\cellcolor{gray!10}\quad\textit{General Baselines:}} \\
\color{gray}Original & -- &\color{gray}{92.6} & \color{gray}{4098} & \color{gray}{53.8} & \color{gray}{12959} & \color{gray}{37.9} & \color{gray}{14576} & \color{gray}{16.7} & \color{gray}{16339} & \color{gray}{2.6} & \color{gray}{20957} & \color{gray}40.7 & \color{gray}13786 \\
Budget Prompt &--& 91.0 & 2979 & 51.7 & 12517 & 39.0 & 13134 & 17.7 & 15768 & 5.1 & 20184 & 40.9 & 12916 \\
SFT-S1.1 & 1K&93.4 & 4101 & 57.7 & 13414 & 39.4 & 15319 & 19.0 & 16572 & 7.7 & 19455 & 43.4 & 13772 \\

\multicolumn{14}{l}{\cellcolor{gray!10}\quad\textit{Specialized Methods:}} \\
EfficientR $\alpha=0.1$ & 3.2K$^\dagger$ &92.2 & 2869 & 51.5 & 11550 & 35.4 & 11902 & 15.4 & 13184 &2.6&18132&39.4&11527 \\
EfficientR $\alpha=0.4$ & 3.2K$^\dagger$ &90.6 & 2432 & 53.5 & 10336 & 33.1 & 11048 & 17.7 & 12216&0.0&18718&39.0&10950\\
AdaptThink & 40K$^\dagger$ & 92.0 & 2095 & 55.0 & 10826 & 36.9 & 12063 & 19.8 & 13873 & 0.0 & 17421& 40.7 & 11256 \\
VeriThinker & 340K & 92.8 & 2674 & 54.6 & 10031 & 34.6 & 10643 & 15.4 & 13689 & 5.1 & 15279 & 40.5 & 10463 \\
\cmidrule[0.4pt](l){1-14}
\textbf{\method} & \textbf{1K} & 92.5 & 1944 & 52.5 & 9899& 37.7 & 9144 & 17.7 & 9629 & 5.1 & 14517 & 40.7 & 9027 \\
\quad \textit{vs. Original} & -- & 
\textcolor{red!70!black}{-0.1} & \textcolor{green!70!black}{-53\%} & 
\textcolor{red!70!black}{-1.3} & \textcolor{green!70!black}{-24\%} & 
\textcolor{red!70!black}{-0.2} & \textcolor{green!70!black}{-37\%} & 
\textcolor{green!70!black}{+1.0} & \textcolor{green!70!black}{-41\%} & 
\textcolor{green!70!black}{+2.5} & \textcolor{green!70!black}{-31\%} &
\textcolor{green!70!black}{+0.4} & \textcolor{green!70!black}{-35\%} \\
\midrule

% ================= DeepSeek 14B =================
\multicolumn{14}{c}{\cellcolor{gray!25}\textit{DeepSeek-R1-Distill-Qwen-14B}} \\ 
\midrule
\color{gray}Original & \color{gray}--&\color{gray}94.1&\color{gray}3800&\color{gray}68.2&\color{gray}11523&\color{gray}46.9&\color{gray}13248&\color{gray}24.8&\color{gray}14916&\color{gray}7.7&\color{gray}19868& \color{gray}48.3 & \color{gray}12671 \\
Budget Prompt & --&93.4 & 2813 & 66.0 & 10513 & 45.6 & 12339 & 24.6 & 13576 & 12.8 & 18541 & 48.5 & 11556 \\
SFT-S1.1 & 1K & 95.3 & 4255 & 69.8 & 12483 & 53.3 & 14104 & 30.4 & 15836 & 10.3 & 20041 & 51.8 & 13344 \\
\cmidrule[0.4pt](l){1-14}
\textbf{\method} & \textbf{1K}&93.4 & 1908 & 63.3 & 7963 & 46.6 & 8430 & 23.8 & 10010 & 12.8 & 15939 & 48.0 & 8850 \\
\quad \textit{vs. Original} & -- & 
\textcolor{red!70!black}{-0.7} & \textcolor{green!70!black}{-50\%} & 
\textcolor{red!70!black}{-4.9} & \textcolor{green!70!black}{-31\%} & 
\textcolor{red!70!black}{-0.3} & \textcolor{green!70!black}{-36\%} & 
\textcolor{red!70!black}{-1.0} & \textcolor{green!70!black}{-33\%} & 
\textcolor{green!70!black}{+5.1} & \textcolor{green!70!black}{-20\%} &
\textcolor{red!70!black}{-0.3} & \textcolor{green!70!black}{-30\%} \\

\midrule

% ================= DeepSeek 32B =================
\multicolumn{14}{c}{\cellcolor{gray!25}\textit{DeepSeek-R1-Distill-Qwen-32B}} \\ 
\midrule
\color{gray}Original &\color{gray}--& \color{gray}94.8 & \color{gray}3611 & \color{gray}69.0 & \color{gray}11061 & \color{gray}51.7 & \color{gray}13051 & \color{gray}26.9 & \color{gray}14813 & \color{gray}15.4 & \color{gray}21151 & \color{gray}51.6 & \color{gray}12737 \\
Budget Prompt & -- & 93.6 & 2481 & 67.7 & 9401 & 47.5 & 11982 & 25.6 & 12463 & 7.7 & 17026 & 48.3 & 10671 \\
SFT-S1.1 & 1K & 96.0 & 3891 & 77.3 & 11383 & 64.8 & 13983 & 32.7 & 15276 & 15.4 & 18836 & 57.2 & 12674 \\
\cmidrule[0.4pt](l){1-14}
\textbf{\method} & \textbf{1K} & 95.9 & 1927 & 68.8 & 8173 & 51.0 & 7908 & 26.2 & 9551 & 12.8 & 18650 & 50.9 & 9242 \\
% \textbf{\method}504 & \textbf{1K} &95.2 & 1945 & 68.8 & 8004 & 47.3 & 7832 & 24.4 & 9650 & 10.3 & 20161 & 49.2 & 9518 \\
\quad \textit{vs. Original} & -- & 
\textcolor{green!70!black}{+1.1} & \textcolor{green!70!black}{-47\%} & 
\textcolor{red!70!black}{-0.2} & \textcolor{green!70!black}{-26\%} &   
\textcolor{red!70!black}{-0.7} & \textcolor{green!70!black}{-39\%} & 
\textcolor{red!70!black}{-0.7} & \textcolor{green!70!black}{-36\%} & 
\textcolor{red!70!black}{-2.6} & \textcolor{green!70!black}{-12\%} &
\textcolor{red!70!black}{-0.7} & \textcolor{green!70!black}{-27\%} \\

\bottomrule
\multicolumn{14}{l}{\footnotesize \textsuperscript{$\dagger$} Online RL methods with 8 rollouts/prompt (EfficientR) and 16 rollouts/prompt (AdaptThink). Unmarked: offline methods.
}

\end{tabular}}
\end{table}

\subsection{Main Results: Mathematical Reasoning}
\label{subsec:main_math}

Table~\ref{tab:main_results} summarizes performance across four model scales. The results demonstrate a consistent trend: \method significantly reduces computational cost while preserving or enhancing reasoning accuracy.
\paragraph{Reasoning Efficiency.}
Across competition-level benchmarks, \method achieves a 24--66\% reduction in token consumption with negligible impact on performance. Notably, DeepSeek-32B on MATH500 yields a 1.1\% accuracy gain while reducing inference tokens by 47\%. This represents a Pareto improvement rather than a trade-off, suggesting that standard Chain-of-Thought (CoT) processes often expend significant computation on redundant reasoning steps which \method successfully prunes.

\paragraph{Data Efficiency.}
\method requires only 1K training samples versus significantly larger data requirements for existing baselines. For SFT-based methods, VeriThinker uses 350K samples and SFT-S1.1 uses 1K samples of higher-cost curated traces. For RL-based methods, the comparison requires accounting for rollout overhead: EfficientR uses 3.2K prompts $\times$ 8 rollouts $= 25.6$K total generations, and AdaptThink uses 40K prompts $\times$ 16 rollouts $= 640$K total generations. Compared on total generations, \method achieves 25.6--640$\times$ greater sample efficiency with no online exploration. Compared on unique prompts alone, \method matches or exceeds efficiency at 3.2--40$\times$ fewer prompts. This demonstrates that adaptive reasoning can emerge from precise data curation rather than exhaustive RL exploration.

\paragraph{Temporal Generalization (AMO-Bench).}
To assess robustness against memorization, we evaluate on 
AMO-Bench, a \textit{temporally held-out} dataset of IMO-level 
problems released after all model training cutoffs. Most model 
variants maintain accuracy within $\pm$3\% of baseline while 
achieving modest token reduction. Qwen3-4B shows a larger 
accuracy gain, which we attribute to the mitigation of two 
overthinking failure modes: the model either contradicts a 
correct intermediate result through excessive reflection, or 
exhausts its context on redundant verification loops. By 
teaching \textit{when to stop reasoning}, \method turns token 
reduction and accuracy improvement into mutually reinforcing 
effects. The lower compression rate on AMO-Bench compared to 
in-distribution math further confirms that the model 
conservatively retains reasoning steps on genuinely novel 
problems, in contrast to SFT-S1.1, which compresses more 
aggressively but at the cost of accuracy.

%------------------------------------------------------------------------------
% TABLE 2: CROSS-DOMAIN GENERALIZATION (Refined)
%------------------------------------------------------------------------------
\begin{table*}[t]
\centering
\caption{Cross-domain generalization results. \method preserves accuracy on out-of-distribution (OOD) tasks. Notably, it does \textbf{not} force compression on unfamiliar domains, providing evidence of a content-aware reasoning policy.}
\label{tab:cross_domain}
\footnotesize
\setlength{\tabcolsep}{3.5pt} % Adjust column spacing

% --- MINIPAGE A: CODE ---
\begin{minipage}{0.48\textwidth}
\centering
\textbf{(a) Code Generation (LiveCodeBench)}
\vspace{1mm}
\resizebox{\linewidth}{!}{
\begin{tabular}{l|cc|cc|cc|cc}
\toprule
\multirow{2}{*}{\textbf{Method}} 
& \multicolumn{2}{c|}{\textbf{Qwen3-4B}} 
& \multicolumn{2}{c|}{\textbf{DS-R1-7B}} 
& \multicolumn{2}{c|}{\textbf{DS-R1-14B}} 
& \multicolumn{2}{c}{\textbf{DS-R1-32B}} \\
\cmidrule(lr){2-3} \cmidrule(lr){4-5} \cmidrule(lr){6-7} \cmidrule(lr){8-9}
& Acc & Tok(k) & Acc & Tok(k) & Acc & Tok(k) & Acc & Tok(k) \\
\midrule
\rowcolor{gray!10} Original & 62.7 & 17.4 & 37.3 & 11.8 & 49.3 & 10.8 & 56.7 & 11.1 \\
Budget Prompt & -- & -- & -- & -- & -- & -- & -- & -- \\
SFT-S1.1 & 50.4 & 14.7 & 39.2 & 15.0 & 57.1 & 12.1 & 65.7 & 11.8 \\
\midrule
\textbf{\method} & \textbf{61.9} & \textbf{13.2} & \textbf{37.7} & \textbf{16.6} & \textbf{53.7} & \textbf{13.8} & \textbf{66.0} & \textbf{13.0} \\
\multicolumn{1}{r|}{\textit{\scriptsize vs.\ Orig}} & 
\scriptsize \textcolor{red!70!black}{-0.8} & \scriptsize \textcolor{green!60!black}{-24\%} & 
\scriptsize \textcolor{green!60!black}{+0.4} & \scriptsize \textcolor{red!70!black}{+41\%} & 
\scriptsize \textcolor{green!60!black}{+4.4} & \scriptsize \textcolor{red!70!black}{+28\%} & 
\scriptsize \textcolor{green!60!black}{+9.3} & \scriptsize \textcolor{red!70!black}{+17\%} \\
\bottomrule
\end{tabular}
}
\end{minipage}
\hfill
% --- MINIPAGE B: KNOWLEDGE ---
\begin{minipage}{0.48\textwidth}
\centering
\textbf{(b) Scientific Knowledge (GPQA-Diamond)}
\vspace{1mm}
\resizebox{\linewidth}{!}{
\begin{tabular}{l|cc|cc|cc|cc}
\toprule
\multirow{2}{*}{\textbf{Method}} 
& \multicolumn{2}{c|}{\textbf{Qwen3-4B}} 
& \multicolumn{2}{c|}{\textbf{DS-R1-7B}} 
& \multicolumn{2}{c|}{\textbf{DS-R1-14B}} 
& \multicolumn{2}{c}{\textbf{DS-R1-32B}} \\
\cmidrule(lr){2-3} \cmidrule(lr){4-5} \cmidrule(lr){6-7} \cmidrule(lr){8-9}
& Acc & Tok(k) & Acc & Tok(k) & Acc & Tok(k) & Acc & Tok(k) \\
\midrule
\rowcolor{gray!10} Original & 66.2 & 8.8 & 53.5 & 8.4 & 60.6 & 7.6 & 60.6 & 7.3 \\
Budget Prompt & 66.2 & 8.7 & 46.5 & 6.5 & 53.0 & 5.5 & 57.1 & 4.9 \\
SFT-S1.1 & 69.2 & 7.6 & 50.5 & 9.6 & 61.1 & 10.0 & 63.6 & 9.3 \\
\midrule
\textbf{\method} & \textbf{67.2} & \textbf{7.7} & \textbf{52.0} & \textbf{10.4} & \textbf{58.6} & \textbf{10.1} & \textbf{63.1} & \textbf{10.5} \\
\multicolumn{1}{r|}{\textit{\scriptsize vs.\ Orig}} & 
\scriptsize \textcolor{green!60!black}{+1.0} & \scriptsize \textcolor{green!60!black}{-13\%} & 
\scriptsize \textcolor{red!70!black}{-1.5} & \scriptsize \textcolor{red!70!black}{+23\%} & 
\scriptsize \textcolor{red!70!black}{-2.0} & \scriptsize \textcolor{red!70!black}{+34\%} & 
\scriptsize \textcolor{green!60!black}{+2.5} & \scriptsize \textcolor{red!70!black}{+44\%} \\
\bottomrule
\end{tabular}
}
\end{minipage}
\end{table*}
%------------------------------------------------------------------------------

\subsection{Cross-Domain Generalization: The Efficiency Spectrum}
\label{subsec:cross_domain}
A critical risk in reasoning compression is degraded performance 
on domains absent from training. We evaluate \method on Code 
Generation and Scientific Reasoning to assess this risk.

\paragraph{Accuracy holds; compression is domain-selective.}
Table~\ref{tab:cross_domain} reveals a consistent asymmetry: 
\method preserves or improves accuracy on both OOD domains, 
while token consumption does not decrease and in some cases 
increases substantially. Rather than compressing indiscriminately, 
the model defaults to fuller reasoning on inputs outside the 
learned compression regime---in contrast to Budget Prompt, which 
forces compression regardless of domain and incurs notable 
accuracy drops on scientific reasoning.

\paragraph{The efficiency spectrum.}
Across benchmarks, compression scales with domain familiarity: 
strongest on in-distribution math, moderate on temporally 
held-out problems, and largely absent on OOD domains. This 
gradient suggests \method internalizes a compression boundary 
tied to its training distribution rather than a uniform 
reasoning policy. We leave a fuller characterization of this 
boundary to future work.

%% ============================================================
%% TABLE 1 — Main Ablation
%% ============================================================
\begin{table}[t]
\centering
\caption{%
  Ablation study on data composition.
}
\label{tab:ablation}
\resizebox{0.65\linewidth}{!}{
\begin{tabular}{l|c|cc|cc}
\toprule
\multirow{2}{*}{\textbf{Method}} & \multirow{2}{*}{\textbf{Count}}
  & \multicolumn{2}{c|}{\textbf{AIME25}}
  & \multicolumn{2}{c}{\textbf{HMMT25}} \\
\cmidrule(lr){3-4}\cmidrule(lr){5-6}
  & & Acc & Tok\,(k) & Acc & Tok\,(k) \\
\midrule
\rowcolor{gray!10}\multicolumn{6}{l}{\textit{Baselines}} \\
Original (zero-shot)             & ---  & 73.3 & 22.0 & 40.0 & 25.8 \\
SFT on external traces           & 1k   & 46.6 & 13.2 & 50.0 & 13.2 \\
\midrule
\rowcolor{blue!5}
\textbf{Ours (full method)}      & \textbf{1k}
  & \textbf{50.0} & \textbf{7.5}
  & \textbf{53.3} & \textbf{15.6} \\
\midrule
\rowcolor{gray!10}\multicolumn{6}{l}{\textit{(a) Answer-correctness filtering}} \\
\quad Correct traces only        & 421  & 43.3 &  8.1 & 33.3 &  9.3 \\
\midrule
\rowcolor{gray!10}\multicolumn{6}{l}{\textit{(b) Trace source}} \\
\quad External teacher (DeepSeek-R1) & 1k & 43.3 & 10.8 & 26.7 & 13.1 \\
\midrule
\rowcolor{gray!10}\multicolumn{6}{l}{\textit{(c) Episode tag format}} \\
\quad Random word tags           & 1k   & 56.6 & 10.6 & 43.3 & 17.6 \\
\quad Special tokens only        & 1k   & 26.7 & 13.1 & 43.3 & 10.8 \\
\bottomrule
\end{tabular}}
\end{table}

\begin{figure}[t]
  \centering
  \includegraphics[width=\linewidth]{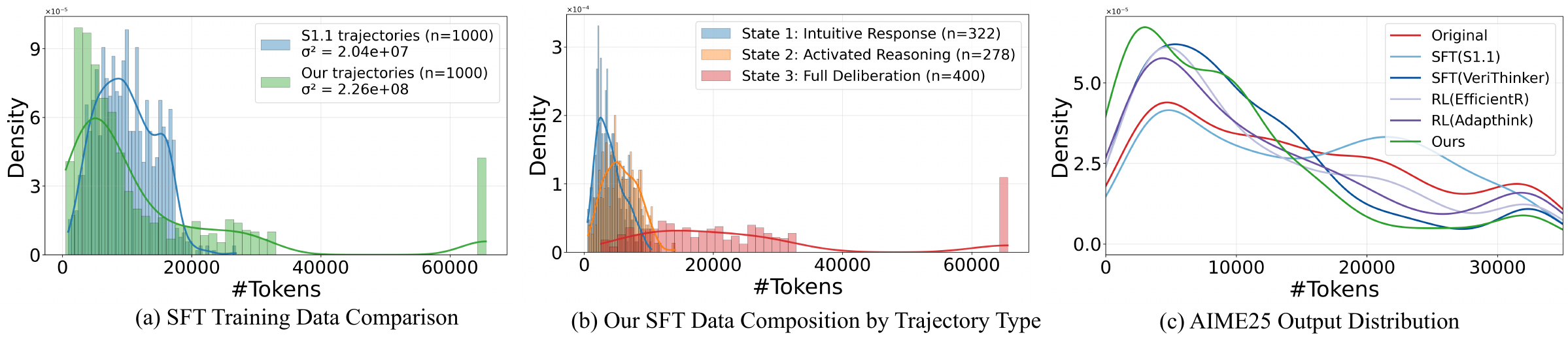}
  \caption{{Training Data Variance and Adaptive CoT Compression.}
  {(a)} Comparison of token length variance between our SFT dataset 
  and the baseline (DPSK trajectories), showing a broader distribution in 
  our method.
  {(b)} Token distribution across three distinct cognitive trajectory 
  types within our dataset.
  {(c)} Post-SFT inference behavior AIME25. The 
  model demonstrates benchmark-specific compression, dynamically adapting 
  reasoning length to problem complexity.}
  \label{fig:distribution}
\end{figure}

%% ============================================================
%% TABLE 2 — Probe ablation (left) | Sample efficiency (right)
%% ============================================================
\begin{table}[t]
\centering
\caption{%
  \textbf{Left}: Probe ablation on R1-Distill-14B (AIME24);
  same-family probes better preserve reasoning alignment.
  \textbf{Right}: Sample efficiency on R1-Distill-7B (AIME24);
  \textit{s1} = base curated set.
  Performance plateaus beyond 1k samples, suggesting trace 
  quality matters more than quantity.
}
\label{tab:probe_sample}
\begin{minipage}{0.48\textwidth}
\centering
\resizebox{\linewidth}{!}{%
\begin{tabular}{llcc}
\toprule
\textbf{Probe} & \textbf{Relationship} & \textbf{Acc} & \textbf{Tok\,(k)} \\
\midrule
\textit{Original}    & ---                 & 70.6 & 11.5 \\
\midrule
Llama-3.1-8B         & cross-family        & 58.1 & 14.5 \\
Qwen3-4B             & same family         & 59.2 &  8.0 \\
Qwen3-30B-A3B        & same family, MoE    & 61.1 &  9.9 \\
Qwen2.5-72B          & same family, larger & 65.2 & 10.5 \\
\bottomrule
\end{tabular}}
\end{minipage}
\hfill
\begin{minipage}{0.48\textwidth}
\centering
\resizebox{\linewidth}{!}{%
\begin{tabular}{rlcc}
\toprule
\textbf{Samples} & \textbf{Source} & \textbf{Acc} & \textbf{Tok\,(k)} \\
\midrule
 200 & \textit{s1} (subsample)        & 50.0 & 16.4 \\
 500 & \textit{s1} (subsample)        & 50.0 & 12.8 \\
 800 & \textit{s1} (subsample)        & 46.7 & 11.2 \\
\textbf{1000} & \textbf{\textit{s1}}  & \textbf{52.5} & \textbf{9.9} \\
2000 & \textit{s1} + MATH/NuminaMath  & 51.6 & 10.2 \\
\bottomrule
\end{tabular}}
\end{minipage}
\end{table}

\section{Analysis}
\label{sec:analysis}

We conduct systematic ablations to understand the mechanisms 
underlying \method, examining four design choices: answer 
correctness filtering, trace source, episode tag format, and 
instruct model selection. Our findings suggest that, for 
reasoning models, the quality of the reasoning process can 
matter more than the correctness of the final outcome.

%% --------------------------------------------------------
\subsection{Process Matters More Than Outcome}
\label{subsec:process_vs_outcome}

\paragraph{Incorrect answers still teach reasoning.}
We compare training on all 1K collected trajectories against 
a filtered subset of 421 correct-only traces. The unfiltered 
set substantially outperforms the filtered subset: 50.0\% 
vs.\ 43.3\% on AIME25 and 53.3\% vs.\ 33.3\% on HMMT25 
(Table~\ref{tab:ablation}, group~(a)). Trajectories with 
incorrect final answers encode valuable reasoning 
patterns---problem decomposition, intermediate verification, 
and explicit self-correction---that teach the model 
\emph{how} to reason rather than \emph{what} to answer.

\paragraph{Reasoning diversity underlies this effect.}
Figure~\ref{fig:distribution}(a) offers a structural 
explanation. The unfiltered set exhibits substantially higher 
variance in trajectory length ($\sigma \approx 36.3$k tokens) 
than the correct-only subset ($\sigma \approx 8.3$k tokens). 
Correct-answer filtering disproportionately removes long, 
exploratory trajectories on difficult problems, discarding 
the reasoning patterns most relevant to hard instances. 
As a result, models trained on the high-variance corpus 
dynamically scale computation to problem difficulty 
(Figure~\ref{fig:distribution}(c)), an adaptive behavior 
largely absent in the correct-only-trained model.

%% --------------------------------------------------------
\subsection{Trace Source and Format}
\label{subsec:trace_source_format}

\paragraph{Self-generated traces outperform external sources.}
We contrast endogenous traces from the target model against 
exogenous traces from a stronger external model (DeepSeek-R1). 
Despite the external model's superior capability, 
self-generated traces yield markedly better outcomes 
(Table~\ref{tab:ablation}, group~(b)), with distributional 
proximity mattering more than raw capability 
(Table~\ref{tab:probe_sample}, left). A stronger model's 
reasoning strategies diverge from the target's generative 
distribution in abstraction level, notation, and granularity, 
forcing simultaneous style adaptation and capability 
improvement. Self-generated traces instead reinforce patterns 
already latent in the model.

\paragraph{Structure matters; semantics do not.}
Replacing semantic tags with arbitrary words yields comparable 
performance, while removing delimiters entirely causes sharp 
degradation (Table~\ref{tab:ablation}, group~(c)). The active 
ingredient is \emph{structural separation} between thinking 
and answering phases, not tag semantics. Consistent with this, 
correctly answered problems consistently use fewer tokens than 
incorrect ones across all models, suggesting compression is a 
byproduct of successful reasoning rather than a cause.
%% --------------------------------------------------------
\subsection{Instruct Model as Difficulty Probe}
\label{subsec:probe_selection}

The instruct probe determines $K^*$ labels and directly 
shapes the training signal. Table~\ref{tab:probe_sample} 
(left) compares probes of varying scale and family on 
R1-Distill-14B.

\paragraph{Same-family probes outperform cross-family ones.}
A cross-family probe (Llama-3.1-8B) yields the largest 
accuracy drop, while same-family probes of increasing scale 
progressively recover performance, consistent with the 
distributional alignment findings in 
Section~\ref{subsec:trace_source_format}.

\paragraph{The probe's failure mode is conservative by design.}
When the probe fails at all hint levels, the sample is 
assigned Full-Hint and the complete reasoning trace is 
preserved. An overly weak probe therefore over-assigns to 
Full-Hint, producing less compression but not incorrect 
supervision. This explains the larger accuracy gap at 14B 
scale with the default Qwen3-4B probe: switching to 
Qwen2.5-72B recovers 6.0 points on AIME24 while retaining 
meaningful token reduction. We recommend using the strongest 
available same-family instruct model as the probe when 
accuracy is the primary concern.
\section{Conclusion}
\label{sec:conclusion}
We introduced \method{}, a data-efficient framework that calibrates
reasoning depth by using the instruct model as a difficulty probe. 
The Minimum Effective Hint criterion converts the abstract question 
of ``how much reasoning is enough'' into a concrete, measurable 
signal, yielding a three-state training set that delivers 24--66\% 
token reduction with only 1K samples. Our results suggest that 
adaptive reasoning can emerge from precise data curation rather 
than exhaustive RL exploration, and that, on problems within the 
training distribution, shorter reasoning and higher accuracy are 
not necessarily at odds---teaching the model \textit{when} to 
stop can mitigate overthinking failure modes.

\paragraph{Limitations and future work.}
\method relies on a capability gap between the instruct and 
reasoning models to generate meaningful difficulty variation; 
how this gap should be calibrated across model families and 
scales remains an open design question. Our experiments focus 
on mathematical reasoning, and whether the learned compression 
boundary generalizes to other domains requires further 
investigation. Finally, the mechanism by which structural 
episode tags activate adaptive reasoning is empirically 
observed but theoretically unexplained.

%==============================================================================
{\small
\bibliographystyle{unsrt}
\bibliography{references}
}
%%%%%%%%%%%%%%%%%%%%%%%%%%%%%%%%%%%%%%%%%%%%%%%%%%%%%%%%%%%%
\newpage

\appendix

%==============================================================================
% IMPLEMENTATION DETAILS
%==============================================================================
\section{Implementation Details}
\label{app:training}

\subsection{Training Configuration}

We perform full-parameter supervised fine-tuning on 8$\times$NVIDIA H20-141GB GPUs, using the Relax framework~\cite{relax2026}. All models are trained with the AdamW optimizer using a cosine learning rate schedule with 10\% warmup for 10 epochs. We set per-device batch size to 1 with gradient accumulation over 8 steps, yielding an effective batch size of 8. The maximum sequence length is 32,768 tokens, and we employ BF16 mixed-precision training with FlashAttention-2.

For smaller models (Qwen3-4B-Think and DeepSeek-R1-Distill-Qwen-7B), we use DeepSpeed ZeRO Stage 0 with sequence parallel size 2 and a learning rate of $1 \times 10^{-5}$. For larger models (DeepSeek-R1-Distill-Qwen-14B and 32B), we adopt DeepSpeed ZeRO Stage 3 to accommodate memory constraints. The 14B model uses learning rate $1 \times 10^{-5}$ and sequence parallel size 2, while the 32B model uses a higher learning rate of $1 \times 10^{-4}$ and sequence parallel size 4.

We evaluate checkpoints on a held-out validation set and select the best-performing one: Qwen3-4B-Think at step 320, DeepSeek-R1-Distill-Qwen-7B at step 450, DeepSeek-R1-Distill-Qwen-14B at step 350, and DeepSeek-R1-Distill-Qwen-32B at step 315.

\subsection{Evaluation Protocol}

\paragraph{Sampling Strategy.}
All evaluations use the \texttt{lighteval} framework\footnote{\url{https://huggingface.co/docs/lighteval}} for reproducibility. For each problem, we sample 16 responses using temperature $T=0.6$, top-$p=0.95$, and maximum new tokens of 32,768. We report averaged accuracy across all samples to mitigate variance inherent in reasoning tasks.

\paragraph{Prompt Format.}
All math problems are evaluated using a consistent prompt format that instructs the model to reason step by step: \textit{``Please reason step by step, and put your final answer within \textbackslash boxed\{\}.''} This ensures fair comparison across different model configurations.

\paragraph{Answer Validation.}
Following the s1 dataset ~\citep{muennighoff2025s1}, we employ an LLM-as-judge approach for answer validation. The grading prompt (Figure~\ref{fig:grading_prompt}) instructs the judge model to compare generated answers against ground-truth solutions, evaluating both final answers and intermediate reasoning steps where applicable.

\begin{figure}[h]
\centering
\fbox{\begin{minipage}{0.93\linewidth}
\small
\textbf{Answer Grading Prompt}

\vspace{0.2cm}
You are an AI assistant for grading a science problem. The user will provide you with the question itself, an attempt made by a student and the correct answer to the problem. Your job is to judge whether the attempt is correct by comparing it with the correct answer. If the expected solution concludes with a number or choice, there should be no ambiguity. If the expected solution involves going through the entire reasoning process, you should judge the attempt based on whether the reasoning process is correct with correct answer if helpful.

The user will provide the attempt and the correct answer in the following format:

\texttt{\# Problem} \\
\texttt{\{problem\}}

\texttt{\#\# Attempt} \\
\texttt{\{attempt\}}

\texttt{\#\# Correct answer} \\
\texttt{\{solution\}}

Explain your reasoning, and end your response on a new line with only ``Yes'' or ``No'' (without quotes).
\end{minipage}}
\caption{Grading prompt for answer validation.}
\label{fig:grading_prompt}
\end{figure}

\paragraph{Token Counting.}
Reasoning overhead is measured by counting tokens within the \texttt{<think>...</think>} tags. Tokens outside these tags (i.e., the problem statement and final answer) are excluded from compression metrics but included when computing total inference cost.

\subsection{Baseline Configuration}
\label{app:baseline}

\paragraph{Budget Forcing Prompt.}
For the Budget Forcing baseline~\citep{han2024token}, we use a token budget of 8,000 tokens and the following prompt template:

\begin{figure}[h]
\centering
\fbox{\begin{minipage}{0.93\linewidth}
\small
\texttt{\{problem\}}

\vspace{0.2cm}
Solve this problem concisely in approximately 8000 tokens.
Show your reasoning with only the essential steps needed.
Put your final answer within \textbackslash boxed\{\}.
\end{minipage}}
\caption{Prompt template for Budget Forcing baseline.}
\label{fig:budget_prompt}
\end{figure}

%==============================================================================

%==============================================================================
% ALGORITHM
%==============================================================================
\section{Algorithm Pseudocode}
\label{app:algorithm}

Algorithm~\ref{alg:pipeline} presents the complete procedure for constructing variable-length Chain-of-Thought training data. The algorithm takes a problem set $\mathcal{X}$, a reasoning model $M_{\text{reason}}$ (e.g., DeepSeek-R1), and an instruct model $M_{\text{instruct}}$ (e.g., Qwen3-4B) as inputs. For each problem, we first generate a complete reasoning trace using the reasoning model, then segment it into episodes, and finally probe the instruct model to determine the minimum hint required for successful problem-solving. Based on the probing result $K^*$, each sample is assigned to one of three cognitive states: \textit{No-Hint} ($K^*=0$), \textit{Sparse-Hint} ($0 < K^* < N$), or \textit{Full-Hint} ($K^*=N$).

\begin{algorithm}[t]
\caption{Variable-Length CoT Data Construction}
\label{alg:pipeline}
\begin{algorithmic}[1]
\REQUIRE Problem set $\mathcal{X}$, Reasoning model $M_{\text{reason}}$, Instruct model $M_{\text{instruct}}$
\ENSURE Training dataset $\mathcal{D} = \mathcal{D}_1 \cup \mathcal{D}_2 \cup \mathcal{D}_3$

\STATE Initialize $\mathcal{D}_1, \mathcal{D}_2, \mathcal{D}_3 \leftarrow \emptyset$

\FOR{\textbf{each} problem $x \in \mathcal{X}$}
    \STATE \textcolor{gray}{\textit{// Stage 1: Generate full reasoning trace}}
    \STATE $r, y \leftarrow M_{\text{reason}}(x)$ \COMMENT{$r$: reasoning trace, $y$: answer}

    \STATE \textcolor{gray}{\textit{// Stage 2: Segment reasoning into episodes}}
    \STATE $(e_1, e_2, \ldots, e_N) \leftarrow \text{SegmentEpisodes}(r)$
    
    \STATE \textcolor{gray}{\textit{// Stage 3: Binary search for minimum effective hint}}
    \STATE $K^* \leftarrow N$ \COMMENT{Default to full reasoning}
    
    \FOR{$k \leftarrow 0$ \textbf{to} $N$}
        \STATE $\hat{y} \leftarrow M_{\text{instruct}}(x \oplus e_{1:k})$ \COMMENT{Probe with $k$ episodes}
        \IF{$\hat{y} = y_{\text{correct}}$}
            \STATE $K^* \leftarrow k$ \COMMENT{Found minimum hint}
            \STATE \textbf{break}
        \ENDIF
    \ENDFOR

    \STATE \textcolor{gray}{\textit{// Stage 4: Assign to cognitive state}}
    \IF{$K^* = 0$}
        \STATE $d \leftarrow (x, \texttt{<think>} \oplus p_0 \oplus \texttt{</think>}, y)$ \COMMENT{No-Hint}
        \STATE $\mathcal{D}_1 \leftarrow \mathcal{D}_1 \cup \{d\}$
    \ELSIF{$0 < K^* < N$}
        \STATE $d \leftarrow (x, \texttt{<think>} \oplus e_{1:K^*} \oplus \texttt{</think>}, y)$ \COMMENT{Sparse-Hint}
        \STATE $\mathcal{D}_2 \leftarrow \mathcal{D}_2 \cup \{d\}$
    \ELSE
        \STATE $d \leftarrow (x, \texttt{<think>} \oplus r \oplus \texttt{</think>}, y)$ \COMMENT{Full-Hint}
        \STATE $\mathcal{D}_3 \leftarrow \mathcal{D}_3 \cup \{d\}$
    \ENDIF
\ENDFOR

\STATE \textbf{return} $\mathcal{D} = \mathcal{D}_1 \cup \mathcal{D}_2 \cup \mathcal{D}_3$
\end{algorithmic}
\end{algorithm}

\paragraph{Episode Segmentation.}
\label{app:segmentation}
The $\text{SegmentEpisodes}(\cdot)$ function splits a reasoning trace into discrete episodes based on reflection keywords that signal cognitive transitions. We identify episode boundaries by detecting the following keywords (case-insensitive): \textit{``wait''}, \textit{``actually''}, \textit{``hmm''}, \textit{``let me reconsider''}, \textit{``on second thought''}, \textit{``hold on''}, and \textit{``let me rethink''}. These keywords naturally demarcate points where the model pauses to reflect, reconsider, or shift its reasoning strategy, making them effective boundaries for incremental hint probing.

% DATA STATISTICS
%==============================================================================
\section{Data Statistics}
\label{app:data}

\subsection{Cognitive State Distribution}

Table~\ref{tab:state_dist} presents the distribution of cognitive states in our constructed dataset after capability probing with Qwen3-4B.

\begin{table}[h]
\centering
\caption{Distribution of cognitive states in the s1K dataset. Think tokens denote reasoning content within \texttt{<think>} tags; answer tokens include all content outside the tags.}
\label{tab:state_dist}
\begin{tabular}{lcccc}
\toprule
\textbf{Cognitive State} & \textbf{Count} & \textbf{Proportion} & \textbf{Think Tokens} & \textbf{Answer Tokens} \\
& & & \textbf{(Median)} & \textbf{(Median)} \\
\midrule
No-Hint ($K^*=0$)       & 357 & 35.7\% & 2      & 3,882 \\
Sparse-Hint ($0<K^*<N$) & 264 & 26.4\% & 215    & 6,522 \\
Full-Hint ($K^*=N$)     & 382 & 38.2\% & 14,894 & 1,034 \\
\midrule
\textbf{Overall} & 1,003 & 100\% & 5,037 & 3,813 \\
\bottomrule
\end{tabular}
\end{table}

\subsection{Token Distribution Analysis}

Several notable patterns emerge from Table~\ref{tab:state_dist}:

\paragraph{Bimodal Reasoning Length Distribution.}
No-Hint samples require virtually no explicit reasoning (median of 2 tokens, corresponding to the directive phrase ``I don't need deep thinking''), while Full-Hint samples demand extensive deliberation (median of 14.9K tokens). Sparse-Hint samples occupy the middle ground (median of 215 tokens), representing genuinely intermediate difficulty levels.

\paragraph{Inverse Correlation with Answer Length.}
Interestingly, harder problems (Full-Hint) tend to produce shorter final answers (median of 1K tokens), whereas easier problems often require more elaborate solution write-ups (median of 3.9K--6.5K tokens). This pattern suggests that the primary challenge lies in \emph{discovering} the solution approach rather than \emph{articulating} the final answer.

\paragraph{Balanced State Distribution.}
The relatively even distribution across states (35.7\% / 26.4\% / 38.2\%) indicates that the s1K dataset encompasses diverse difficulty levels, preventing the model from converging to a single dominant reasoning strategy.

%==============================================================================
% CASE STUDIES
%==============================================================================
\section{Case Studies}
\label{app:cases}

This section presents representative examples from each cognitive state, illustrating how our method adaptively calibrates reasoning depth according to problem difficulty.

\subsection{No-Hint Example (State I)}

Table~\ref{tab:case_nohint} demonstrates a straightforward physics problem that the instruct model can solve without any reasoning hints. The original reasoning trace spans thousands of tokens, but our method compresses this to just 2 tokens—a 99.96\% reduction—while preserving correctness.

\begin{table}[h]
\centering
\caption{No-Hint example: A standard calculation problem solvable via direct computation.}
\label{tab:case_nohint}
\begin{tabular}{p{0.95\linewidth}}
\toprule
\textbf{Problem} \\
Compute the mean molecular speed $v$ in the heavy gas radon (Rn) in m/s. \\
\midrule
\textbf{Original Reasoning Model Output} \\
\texttt{<think>} \\
I may need some deep thinking. Okay, so I need to compute the mean molecular speed of radon gas. The formula for mean molecular speed is $v = \sqrt{\frac{8RT}{\pi M}}$ where $R$ is the gas constant, $T$ is temperature, and $M$ is molar mass... \\
\textit{[...5,037 tokens of detailed derivation...]} \\
\texttt{</think>} \\
The mean molecular speed is approximately $\boxed{169}$ m/s. \\
\midrule
\textbf{Our Method Output} \\
\texttt{<think>} \\
I don't need deep thinking. \\
\texttt{</think>} \\
The mean molecular speed is approximately $\boxed{169}$ m/s. \\
\midrule
\textbf{Analysis:} Compression ratio of 99.96\%. The instruct model possesses sufficient domain knowledge to solve this problem directly without explicit reasoning scaffolding. \\
\bottomrule
\end{tabular}
\end{table}

\subsection{Sparse-Hint Example (State II)}

Table~\ref{tab:case_sparse} shows a problem of moderate difficulty where a partial reasoning hint suffices to activate the model's latent problem-solving capability.

\begin{table}[h]
\centering
\caption{Sparse-Hint example: A problem requiring a key conceptual insight to unlock the solution.}
\label{tab:case_sparse}
\begin{tabular}{p{0.95\linewidth}}
\toprule
\textbf{Problem} \\
\textit{(Moderate difficulty mathematical reasoning problem)} \\
\midrule
\textbf{Original Reasoning Model Output} \\
\texttt{<think>} \\
\textit{[...Full reasoning trace with multiple exploration paths...]} \\
\texttt{</think>} \\
\textit{(Final answer)} \\
\midrule
\textbf{Our Method Output} \\
\texttt{<think>} \\
I may need some thinking. The key insight here is to recognize that... \\
\textit{[...215 tokens capturing essential reasoning steps...]} \\
\texttt{</think>} \\
\textit{(Final answer)} \\
\midrule
\textbf{Analysis:} Compression ratio of 95.7\%. Only the critical reasoning steps are retained, filtering out exploratory tangents and redundant verification loops. \\
\bottomrule
\end{tabular}
\end{table}

\subsection{Full-Hint Example (State III)}

Table~\ref{tab:case_full} presents a challenging combinatorics problem that exceeds the instruct model's independent solving capability, necessitating the complete reasoning trace.

\begin{table}[h]
\centering
\caption{Full-Hint example: A complex problem requiring comprehensive step-by-step reasoning.}
\label{tab:case_full}
\begin{tabular}{p{0.95\linewidth}}
\toprule
\textbf{Problem} \\
Given a rational number, write it as a fraction in lowest terms and calculate the product of the resulting numerator and denominator. For how many rational numbers between 0 and 1 will $20!$ be the resulting product? \\
\midrule
\textbf{Our Method Output} \\
\texttt{<think>} \\
This is a complex question, and it is difficult to provide a direct answer. I need to think deeply about it. \\

Let me think. We are given a problem about rational numbers between 0 and 1 in lowest terms where numerator $\times$ denominator $= 20!$... \\

For a fraction $\frac{a}{b}$ in lowest terms, we need $\gcd(a,b) = 1$ and $a \cdot b = 20!$... \\

Wait, let me verify this approach. Since $a < b$ (as the fraction is between 0 and 1), we need to count coprime pairs... \\

\textit{[...14,894 tokens of detailed mathematical reasoning...]} \\

Therefore, the number of such rational numbers equals $2^7 = 128$. \\
\texttt{</think>} \\
The answer is $\boxed{128}$. \\
\midrule
\textbf{Analysis:} No compression applied. This problem requires sophisticated number-theoretic reasoning that lies beyond the instruct model's independent capability, justifying retention of the complete reasoning trace. \\
\bottomrule
\end{tabular}
\end{table}

\subsection{Compression Summary}

Table~\ref{tab:compression_summary} summarizes the token compression achieved across the three cognitive states, demonstrating the effectiveness of our adaptive approach.

\begin{table}[h]
\centering
\caption{Token compression statistics by cognitive state.}
\label{tab:compression_summary}
\begin{tabular}{lccl}
\toprule
\textbf{State} & \textbf{Median Tokens} & \textbf{Compression} & \textbf{Interpretation} \\
\midrule
No-Hint & 2 & 99.96\% & Direct solution; reasoning unnecessary \\
Sparse-Hint & 215 & 95.7\% & Key insight activates latent capability \\
Full-Hint & 14,894 & 0\% & Complete deliberation required \\
\midrule
\textbf{Weighted Average} & 5,037 & 31.5\% & Adaptive depth calibration \\
\bottomrule
\end{tabular}
\end{table}

%%%%%%%%%%%%%%%%%%%%%%%%%%%%%%%%%%%%%%%%%%%%%%%%%%%%%%%%%%%%

\newpage

\end{document}